\documentclass[letterpaper, 10pt, conference]{ieeeconf}
\IEEEoverridecommandlockouts

\usepackage{cite}
\usepackage{hyperref}
\usepackage{amsmath,amssymb,amsfonts}
\usepackage{algorithmic}
\usepackage{graphicx}
\usepackage{textcomp}
\usepackage{xcolor}
\usepackage{diagbox}
\usepackage{tabularx}
\usepackage{array}
\usepackage{subcaption}
\usepackage{multirow}
\usepackage[utf8]{inputenc}
\usepackage{multirow}
\usepackage{tabularray}
\usepackage{float}
\usepackage{vcell}
\usepackage{cancel}
\usepackage[export]{adjustbox}
\def\BibTeX{{\rm B\kern-.05em{\sc i\kern-.025em b}\kern-.08em
    T\kern-.1667em\lower.7ex\hbox{E}\kern-.125emX}}

\title{LinguaSim: Interactive Multi-Vehicle Testing Scenario Generation via Natural Language Instruction Based on Large Language Models\\
\thanks{This research was funded by National Key R\&D Program of China (2023YFB2504402).}
}

\author{Qingyuan Shi, Qingwen Meng, Hao Cheng, Qing Xu, Jianqiang Wang*}

\begin{document}

\maketitle

\begin{abstract}
The generation of testing and training scenarios for autonomous vehicles has drawn significant attention. While Large Language Models (LLMs) have enabled new scenario generation methods, current methods struggle to balance command adherence accuracy with the realism of real-world driving environments. To reduce scenario description complexity, these methods often compromise realism by limiting scenarios to 2D, or open-loop simulations where background vehicles follow predefined, non-interactive behaviors. We propose \textit{LinguaSim}, an LLM-based framework that converts natural language into realistic, interactive 3D scenarios, ensuring both dynamic vehicle interactions and faithful alignment between the input descriptions and the generated scenarios. A feedback calibration module further refines the generation precision, improving fidelity to user intent. By bridging the gap between natural language and closed-loop, interactive simulations, \textit{LinguaSim} constrains adversarial vehicle behaviors using both the scenario description and the autonomous driving model guiding them. This framework facilitates the creation of high-fidelity scenarios that enhance safety testing and training. Experiments show \textit{LinguaSim} can generate scenarios with varying criticality aligned with different natural language descriptions (ACT: 0.072 s for dangerous vs. 3.532 s for safe descriptions; comfortability: 0.654 vs. 0.764),  and its refinement module effectively reduces excessive aggressiveness in \textit{LinguaSim}'s initial outputs, lowering the crash rate from 46.9\% to 6.3\% to better match user intentions.
\end{abstract}

\medskip


\section{Introduction}

Currently, the systematic generation of scenarios for testing and training autonomous vehicles remains a problematic matter in the field. The most classic and straightforward way is testing vehicles in the real world, which often introduces unavoidable risks and a long testing cycle. Therefore, efforts have been made to generate safety-critical scenarios within specific simulation environments, such as CARLA, SUMO, and DYNA4, aiming to significantly decrease the costs associated with acquiring training and testing data.

\begin{figure}[ht]
    \centering
    \includegraphics[width=\linewidth]{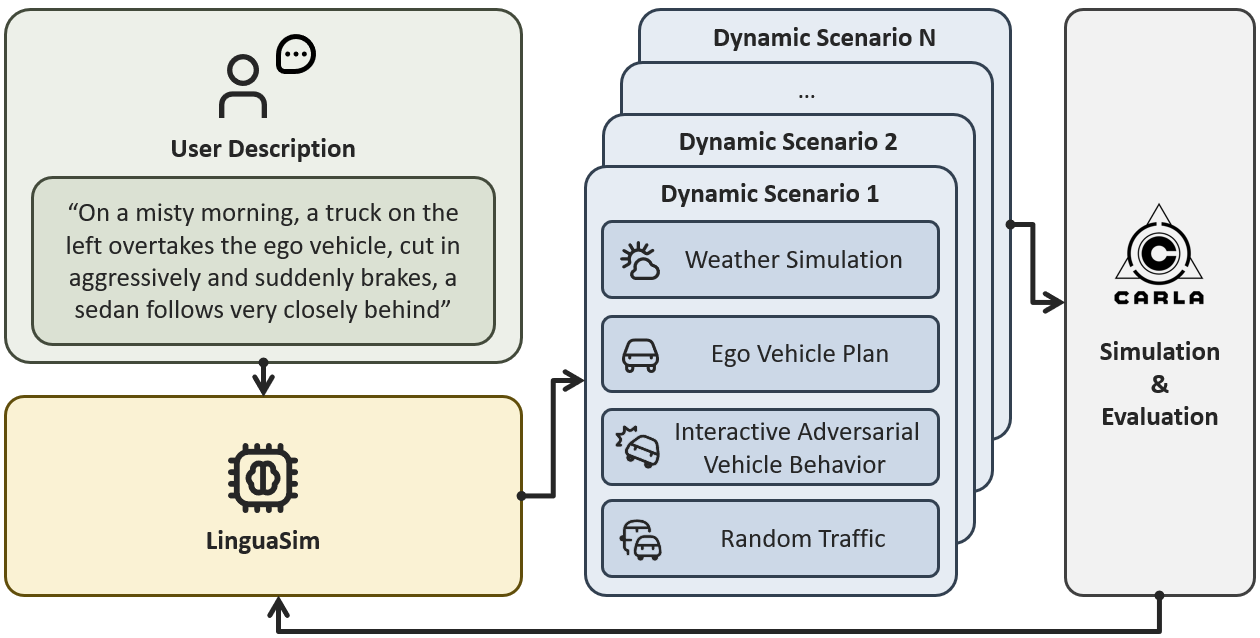}
    \caption{\textit{LinguaSim}: a scenario generation framework based on LLM for autonomous vehicle testing and training}
    \label{fig:linguasim}
\end{figure}

With the rapid development of LLMs, the generation of safety-critical scenarios based solely on user descriptions has become possible. With the help of LLMs, researchers can now generate specific safety scenarios in large quantities, such as generating highway lane-changing scenarios in batches, which can significantly facilitate the testing process to evaluate the performance of the tested vehicle under certain traffic scenes. Multiple research studies have already implemented the bridge between natural language description and simulated scenarios, generating different types of training and testing scenarios for autonomous vehicles \cite{b1, b2, b3, b4, b5}. 


However, one fundamental problem remains: the background vehicles ought to be guided by their integrated autonomous driving model to maintain their own intelligence, and at the same time, the natural language should be respected as well. Taking both of the constraints into account poses a significant challenge. Most current approaches leveraging LLMs generate only open-loop scenarios, where the adversarial vehicles in the scenarios perform fixed behaviors without integrated intelligence, interacting neither with other vehicles nor with the environment in which the scenario takes place \cite{b1}\cite{b2}\cite{b4}. The approach proposed by Lu \textit{et al.} uses a data-driven model to retain the intelligence of the adversarial vehicles, but on the other hand, the adversarial behaviors described in the natural language input are therefore neglected \cite{b3}.

\medskip

To tackle the challenge of generating realistic safety-critical scenarios while respecting natural language descriptions, we propose \textit{LinguaSim}, a novel scenario generation framework. As depicted in Fig.\ref{fig:linguasim}, \textit{LinguaSim} leverages LLMs to transform user descriptions into simulated scenarios. It ensures that all specified elements are guided by certain autonomous driving models while adhering to the natural language descriptions. This approach enables the creation of diverse scenarios without relying on real-world data, thereby enhancing flexibility and realism in testing and training environments.

\medskip

The contribution of this article can be concluded as: 
\begin{itemize}
    \item Proposed \textit{LinguaSim}, a testing scenario generation framework for automated vehicles, transforming natural language input to realistic scenarios. The actors are guided by various autonomous driving models, capable of interacting with the ego vehicle or other elements in the environment,  while following the instructions of the natural language input as the same time.
    \item A layered scenario generation structure is implemented. \textit{LinguaSim} generates different layers of the scenario sequentially, which include the general environment layer, the ego vehicle layer, the adversarial vehicle layer, and the background traffic layer. This four-layered structure splits traffic scenarios into distinct elements that can be generated by different LLM agents, ensuring both the realism and precision of the generated scenario in relation to the natural language input.
    \item A real-time evaluation mechanism is implemented to track criticality, comfortability, and other information frame-by-frame in the scenario. Given the output of the real-time evaluation, the refinement module integrated in \textit{LinguaSim} iteratively aligns the generated scenario with the natural language input, further improving the precision of the generation in relation to the natural language input.
\end{itemize}

\medskip

The following content of this paper will be organized as follows: in Section \ref{sec:related_works}, multiple state-of-the-art methods of test scenario generation are introduced, with their advantages as well as drawbacks briefly reviewed and compared with \textit{LinguaSim}, the scenario generation framework that is about to be proposed. In Section \ref{sec:method} the design pipeline and detailed modular implementation method will be further elaborated. In Section \ref{sec:experiments}, testing scenarios generated directly by \textit{LinguaSim} will be presented, as well as the results of the refinement module. Section \ref{sec:conclusion} will conclude its contribution to the field and potential limitations. 

\begin{figure*}[h]
    \centering
    \includegraphics[width=1\linewidth]{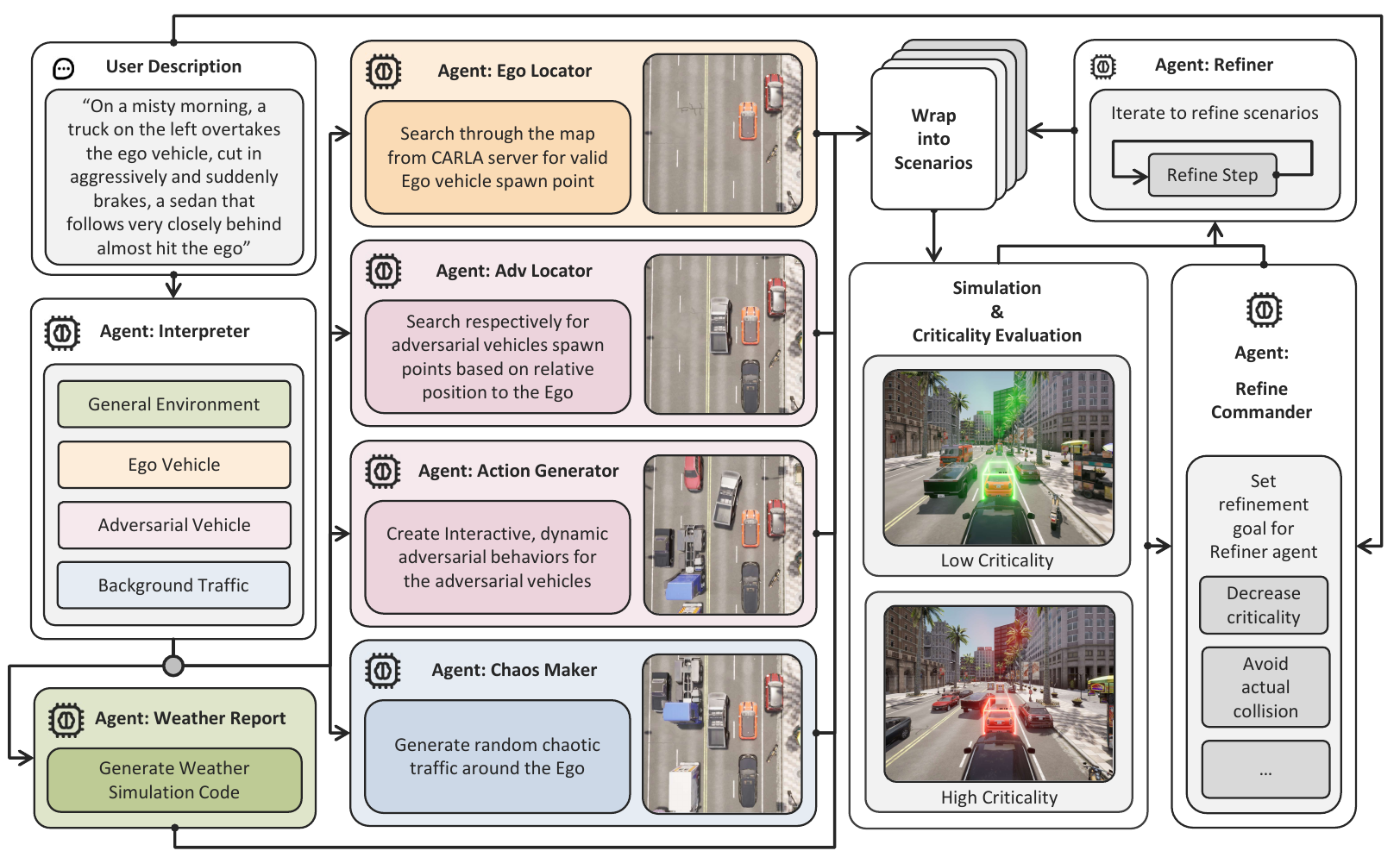}
    \caption{A Overview of The Modular Design of \textit{LinguaSim}}
    \label{fig:pipeline}
\end{figure*}

\section{Related Works}
\label{sec:related_works} 

\subsection{Scenario Generation Leveraging the Power of LLM}
\label{subsec:LLM_generation}
Recent advancements in LLMs have opened new avenues for generating safety-critical scenarios for autonomous vehicles. Current LLM-based frameworks for scenario generation typically follow a structured workflow: they interpret natural language user input and elaborate it into comprehensive scenario descriptions, then systematically derive critical scenario components—including background vehicle behaviors, road geometry specifications, and initial states of surrounding vehicles. These components are generated through specialized LLM agents, each designed to address specific aspects of the scenario. The outputs from these agents are subsequently integrated into a cohesive scenario description that can be executed in simulation environments for comprehensive testing and validation.

\medskip

For instance, ChatScene proposed by Zhang \textit{et al.} utilizes LLM to transform unstructured language instructions into detailed traffic scenarios by first generating textual descriptions, which are then parsed into specific sub-descriptions that define vehicle behaviors and locations \cite{b1}. This process enables the creation of executable code for simulations in the CARLA environment, significantly enhancing the diversity and complexity of generated scenarios. Similarly, Li \textit{et al.} proposed ChatSUMO that integrates LLM capabilities to automate traffic scenario generation within the SUMO simulation environment, allowing users to create customized simulations without extensive traffic simulation expertise \cite{b2}. By converting user inputs into relevant keywords and executing Python scripts, it eventually generates both abstract and real-world traffic scenarios. Furthermore, a multi-modal LLM-based framework, AutoScenario, has been developed by Lu \textit{et al.} to generate realistic corner cases by leveraging diverse real-world data sources \cite{b3}. This framework not only captures key risk factors but also integrates tools from SUMO and CARLA to ensure that the generated scenarios reflect real-world complexities. While these approaches showcase the potential of LLMs in scenario generation, they also face multiple challenges. SUMO-based methods may lack realism for generating 2D-only scenarios; CARLA-based methods suffer from the complexity of the simulation platform. When taking natural language descriptions as input, the previous methods either lose control of the NPC vehicle behavior when using a learnable method to take over completely, or lose the capability to integrate modern adversarial algorithms introduced in Section \ref{subsec:adversarial_ai} while using high-level languages to describe the scenario, such as Scenic and OpenScenario.

\subsection{Scenario Generation Based on Adversarial Agents}
\label{subsec:adversarial_ai}
One of the trending state-of-the-art approaches for scenario generation is to train adversarial models for background vehicles to perform aggressive behaviors. With these adversarial agents serving as background agents, the realism of the scenarios is thus improved thanks to the interaction they bring. Multiple adversarial algorithms have been developed based on this technical approach \cite{b6} \cite{b7}. Hao \textit{et al.} proposed an approach that combines human driving priors with reinforcement learning techniques to create realistic adversarial behaviors for vehicles in the generated scenario \cite{b8}. Furthermore, Chen \textit{et al.} proposed FREA, generating NPCs with a reasonable level of adversariality, preventing unavoidable collisions by evaluating the feasibility of the NPC behaviors \cite{b9}. Those methods assure a realistic and natural simulation by constraining the vehicle behavior via the trained model, as well as keeping the risk level at a high or at least a reasonable level. However, the technical route presented above might lack diversity in generated scenarios for a fixed adversarial algorithm with a fixed, manually set scenario.  

\medskip

Our work, on the other hand, by wrapping adversarial models into \textit{Atomic Behaviors} and using LLM agents to configure and construct them into a \textit{Behavior Topology Web}(will be further explained in section \ref{subsec:action_generation}), successfully combines the advantages of the adversarial algorithm-based methods in Section \ref{subsec:adversarial_ai} and the LLM-based scenario generation method in \ref{subsec:LLM_generation}. This approach allows for huge diversity in the generated scenarios based on a single natural language description input, with the behaviors of the background vehicles constrained by both adversarial vehicle models and the natural language input from the user. 

\section{Method}
\label{sec:method}
In this section, we focus on the generation pipeline of \textit{LinguaSim}, as illustrated in Fig. \ref{fig:pipeline}, highlighting how the tool was developed as separate modules and the unique advantages offered by this structure.
\smallskip

Among similar scenario generation tools based on user natural language descriptions, utilizing the CARLA simulator is generally considered more challenging than using 2D simulators like SUMO, primarily due to the complexity of implementing vehicle control. However, higher-level scenario programming languages such as Scenic and OpenScenario often compromise the controllability of individual traffic members. In contrast, without these high-level languages, the CARLA environment lacks predefined scenarios, which we address in Section \ref{subsec:pipeline}. Additionally, there is no existing method to project natural language directly into vehicle control, a gap we fill with our novel approach in Section \ref{subsec:action_generation}.

\subsection{Pipeline}
\label{subsec:pipeline}
In this section, we present the general pipeline of \textit{LinguaSim} shown in Fig. \ref{fig:pipeline}. A testing scenario could be deconstructed into four layers: 1. general environment layer, 2. ego vehicle layer, 3. adversarial vehicle layer, and 4. background traffic layer, which will be generated by different LLM-based agents sequentially. The deconstruction process itself is handled by LLM agent \textit{Interpreter} on the left side of Fig. \ref{fig:pipeline}, which takes the user description as input, elaborates, and decomposes it into 4 aspects which respectively correspond to the 4-layer structure. \textit{Interpreter} agent will improvise if the information of any layer is absent in the original description, then passes the elaborated description to the following generation layers, which are listed below: 
\begin{enumerate}
    \item \textbf{General Environment Layer}: Corresponding to the green modules in Fig. \ref{fig:pipeline}. This layer mainly contains map and weather information, \textit{LinguaSim} loads the corresponding map, and simulates weather via the LLM agent \textit{Weather Report}, which takes the \textit{General Environment} information from \textit{Interpreter} and outputs the weather configurations. The weather condition could impact the sensor data if perception-involved training or testing are conducted. 
    \item \textbf{Ego Vehicle Layer:} Corresponding to the orange modules in Fig. \ref{fig:pipeline}. This layer contains the road geometry information where the scenario takes place, for instance, in front of a crossroad, a roundabout, or on a straight lane. This layer is handled by LLM agent  \textit{Ego Locator}. \textit{LinguaSim} systematically searches for valid spawn points for the ego vehicle across the map loaded in \textit{General Environment Layer}, providing different environments for generated scenarios.
    \item \textbf{Adversarial Vehicle Layer:} Corresponding to the lavender modules in Fig. \ref{fig:pipeline}. This layer contains the information of the adversarial vehicles that are guided by \textit{LinguaSim} scenarios. The static spawn positions of these vehicles are handled by LLM agent \textit{Adv Locator}, which generates their relative positions to the ego vehicle (e.g., on the left side, behind, or across the same crossroad). \textit{Adv Locator} ensures that both the ego vehicle and adversarial vehicles can be legally placed on the lanes. Furthermore, the dynamic behaviors of these vehicles are managed by LLM agent \textit{Action Generator}, which generates the behavior topology for the scenario and applies control to the target vehicles. This process will be further explained in Section \ref{subsec:action_generation}.
    \item \textbf{Background Traffic Layer:} Corresponding to the blue modules in Fig. \ref{fig:pipeline}. This layer contains the information of the background adversarial vehicles whose behaviors are not directly guided by \textit{LinguaSim}. These vehicles are automatically generated and placed around the ego vehicle and the guided adversarial vehicles by LLM agent \textit{Chaos Maker}, and roam aimlessly on the given map. The background vehicles significantly increase the uncertainty and complexity of the generated scenarios.
\end{enumerate}

\subsection{Adversarial Behavior Generation}
\label{subsec:action_generation}
\begin{figure}[h]
    \centering
    \includegraphics[width=1\linewidth]{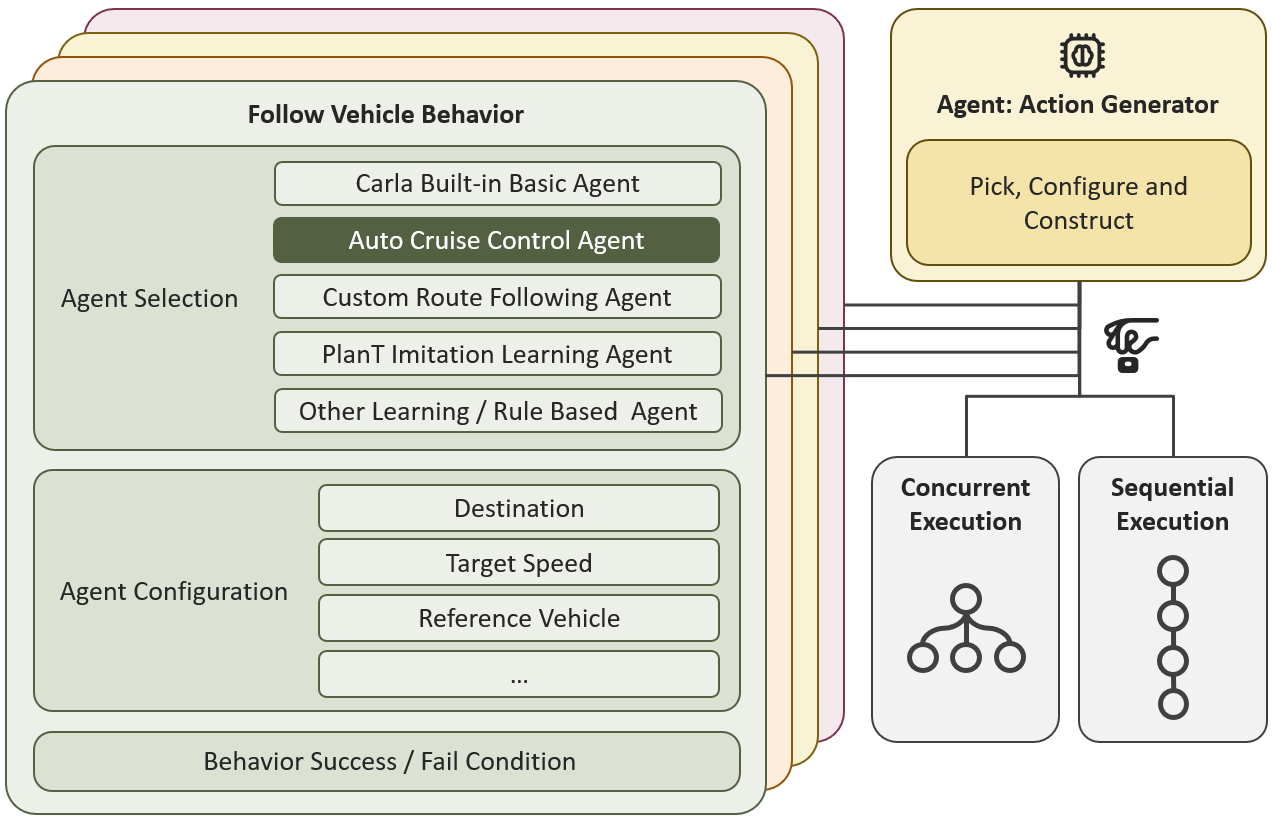}
    \caption{The basic workflow of module \textit{Action Generator}}
    \label{fig:action_generator}
\end{figure}

Compared to other state-of-the-art methods for generating 3D realistic scenarios from natural language descriptions, \textit{LinguaSim} achieves a higher level of realism, flexibility, and interactivity due to the innovative structure of its \textit{Action Generator} agent. The detailed workflow of this component will be elaborated further in this section, with a simplified operational logic of the \textit{Action Generator} illustrated in Fig. \ref{fig:action_generator}.

To establish a solid foundation for the \textit{Action Generator}, a retrieval database was constructed to store various behaviors available for the guided adversarial vehicles. Each behavior in the database is referred to as an \textit{Atomic Behavior}, serving as a fundamental component in the subsequent process. As illustrated in Fig. \ref{fig:action_generator}, each \textit{Atomic Behavior} comprises three essential parts:

\begin{figure}[h]
    \centering
    \includegraphics[width=\linewidth]{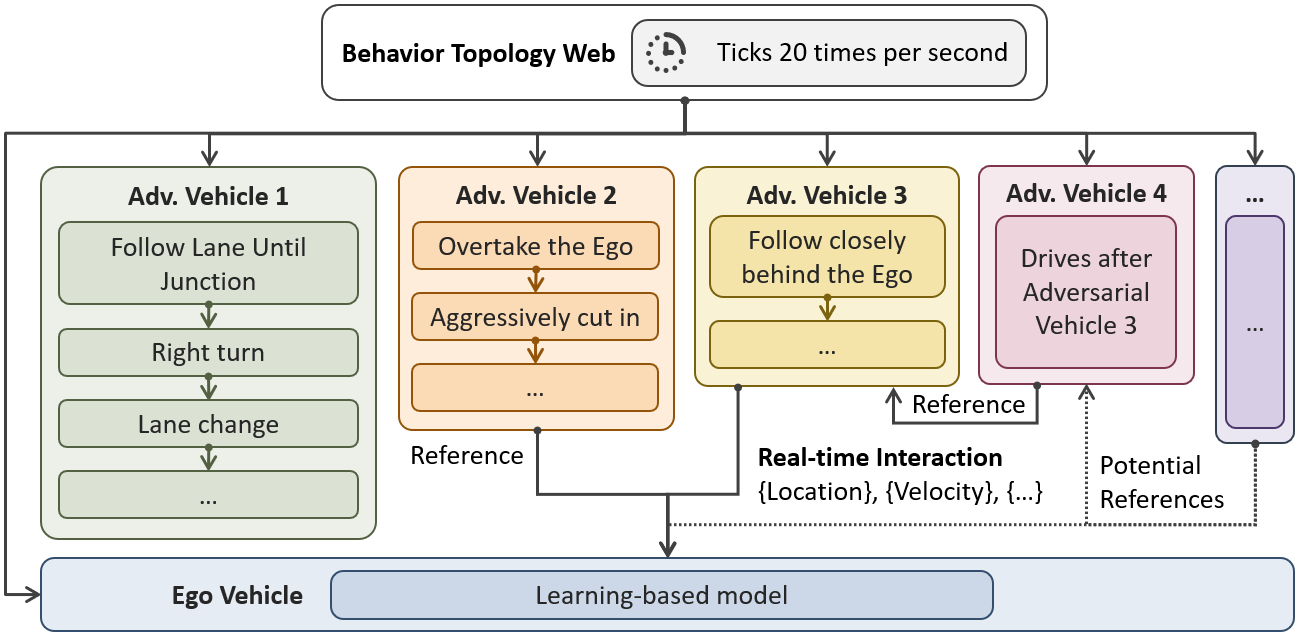}
    \caption{An example of the \textit{Behavior Topology Web} generated by the \textit{Action Generator}}
    \label{fig:behavior_web}
\end{figure}

\begin{enumerate}
\item \textbf{Agent Selection:} An autonomous driving agent is selected to guide the adversarial vehicle to which the \textit{Atomic Behavior} is applied. \textit{LinguaSim} includes various predefined agents, such as the basic CARLA built-in agent that follows a given route, an auto cruise control (ACC) agent that follows the vehicle in front, or the PlanT agent, an imitation-learning-based planning algorithm developed by Renz, Chitta \textit{et al.} \cite{b10}. These agents serve different purposes; for example, the \textit{Follow Vehicle} behavior uses the ACC agent, while the PlanT agent is often used for less aggressive behaviors to mimic cautious drivers.
\item \textbf{Agent Configuration:} The agent provides a behavior pattern for the guided vehicle but requires configuration to execute the desired maneuver. For instance, the \textit{Follow Vehicle} behavior involves configurations such as criteria for determining which vehicle to follow, the target speed for auto cruise control, and the aggressiveness of the following behavior.
\item \textbf{Behavior Success/Fail Condition (Optional):} This specifies the conditions under which the behavior's running state is set to success or fail. For example, the \textit{Stop Vehicle} behavior is considered successful when the vehicle's speed reaches 0, while the \textit{Cut In} behavior is successful if the guided vehicle is in the same lane as the vehicle being cut in.
\end{enumerate}

The \textit{Action Generator} agent selects suitable \textit{Atomic Behaviors}, configures them according to specific requirements, and connects them using either \textit{Concurrent Execution} or \textit{Sequential Execution}, referencing other vehicles during the configuration process. This ultimately constructs an interactive behavior topology web (as illustrated in Fig. \ref{fig:behavior_web}), which operates at a frequency of 20 Hz during the execution of CARLA scenarios.
\medskip

\subsection{Criticality Evaluation and Scenario Refinement}
\label{subsec:criticality_evaluation}
\begin{figure}[h]
    \centering
    \includegraphics[width=1\linewidth]{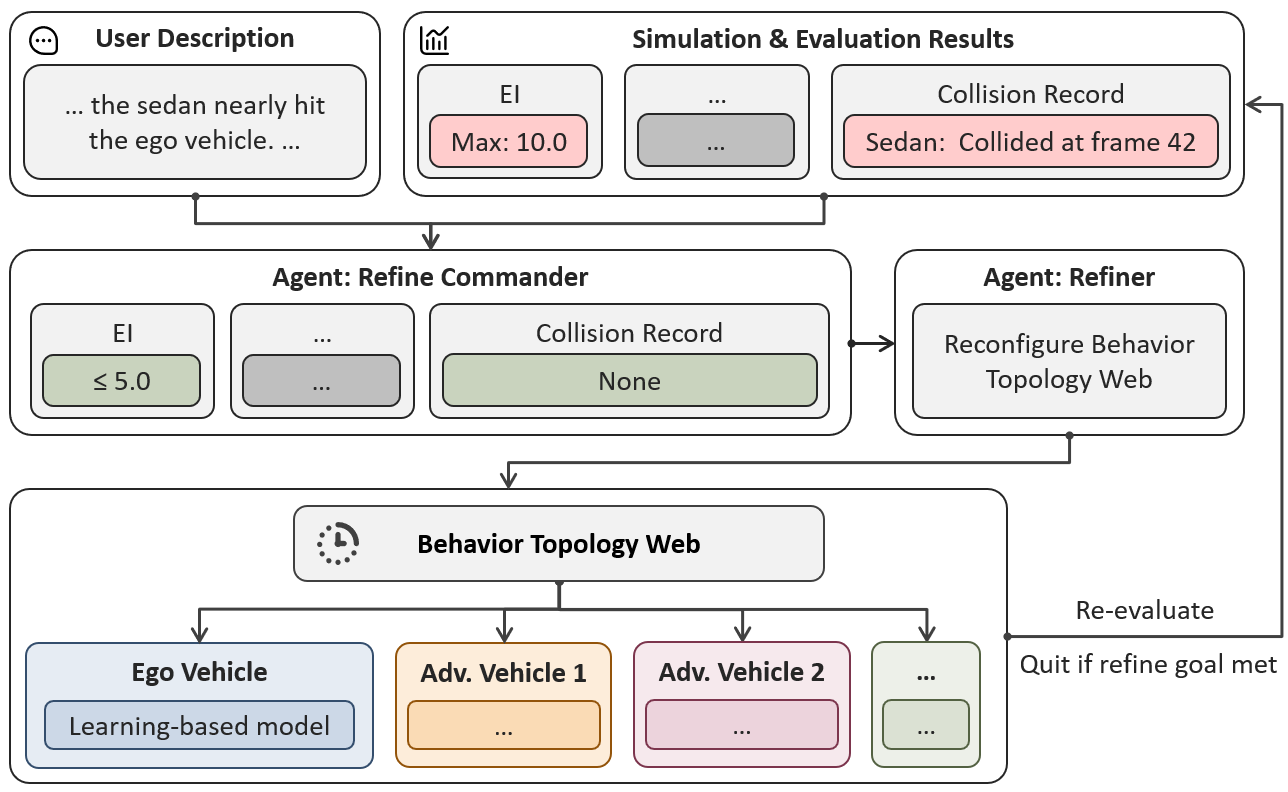}
    \caption{The Basic Pipeline of \textit{Refiner} and \textit{Refine Commander}}
    \label{fig:refiner}
\end{figure}

LLMs can accurately comprehend text within a given text context, but they sometimes struggle to grasp temporal-spatial relationships within a 3D simulation space. To ensure that the generated scenarios align with user demands, such as the "near miss" requirement shown in the user description block in Fig. \ref{fig:pipeline}, where the scenario must be dangerous but without actual collisions, the generated scenarios need to be simulated, evaluated, and then refined by the \textit{Refine Commander} and \textit{Refiner} agents, whose basic workflow is demonstrated in Fig. \ref{fig:refiner}.

\smallskip
For scenario evaluation, we selected multiple metrics to assess the overall or real-time properties of the scenario, including not only basic physical quantities such as velocity, acceleration, and jerk, but also criticality and comfort. Given the paper's space constraints, we prioritized the following two metrics that directly support our objective of evaluating scenario safety and alignment with user descriptions, though researchers can substitute these with alternatives based on their specific evaluation needs.
\begin{itemize}
    \item \textit{Emergency Index (EI)}, proposed by Cheng \textit{et al.}, a surrogate safety measure that quantifies the intensity of evasive actions needed to prevent a crash \cite{b11}. This metric is visually represented in the simulator as the color of the vehicle bounding box, shown in Fig. \ref{fig:pipeline}, where warmer colors indicate higher criticality.
    \item \textit{Collision Record}, a collision sensor integrated into \textit{LinguaSim}, which tracks whether the ego vehicle experiences a collision during scenario playback and identifies the involved parties if a collision occurs.
\end{itemize}

\smallskip
The evaluation results, along with the original user description, are then provided to the \textit{Refine Commander} to assess whether the scenario's criticality aligns with the description. If it does not align, the \textit{Refine Commander} generates a refinement goal and sends it to the \textit{Refiner} for an iterative refinement process that modifies the scenario files. If it does, the \textit{Refiner} will not be invoked.

\section{Experiments}
\label{sec:experiments}

\subsection{Interactive Scenario Generation}
\begin{figure}[htbp]
    \centering
    \begin{subfigure}[b]{0.155\textwidth}
        \centering
        \includegraphics[width=\textwidth]{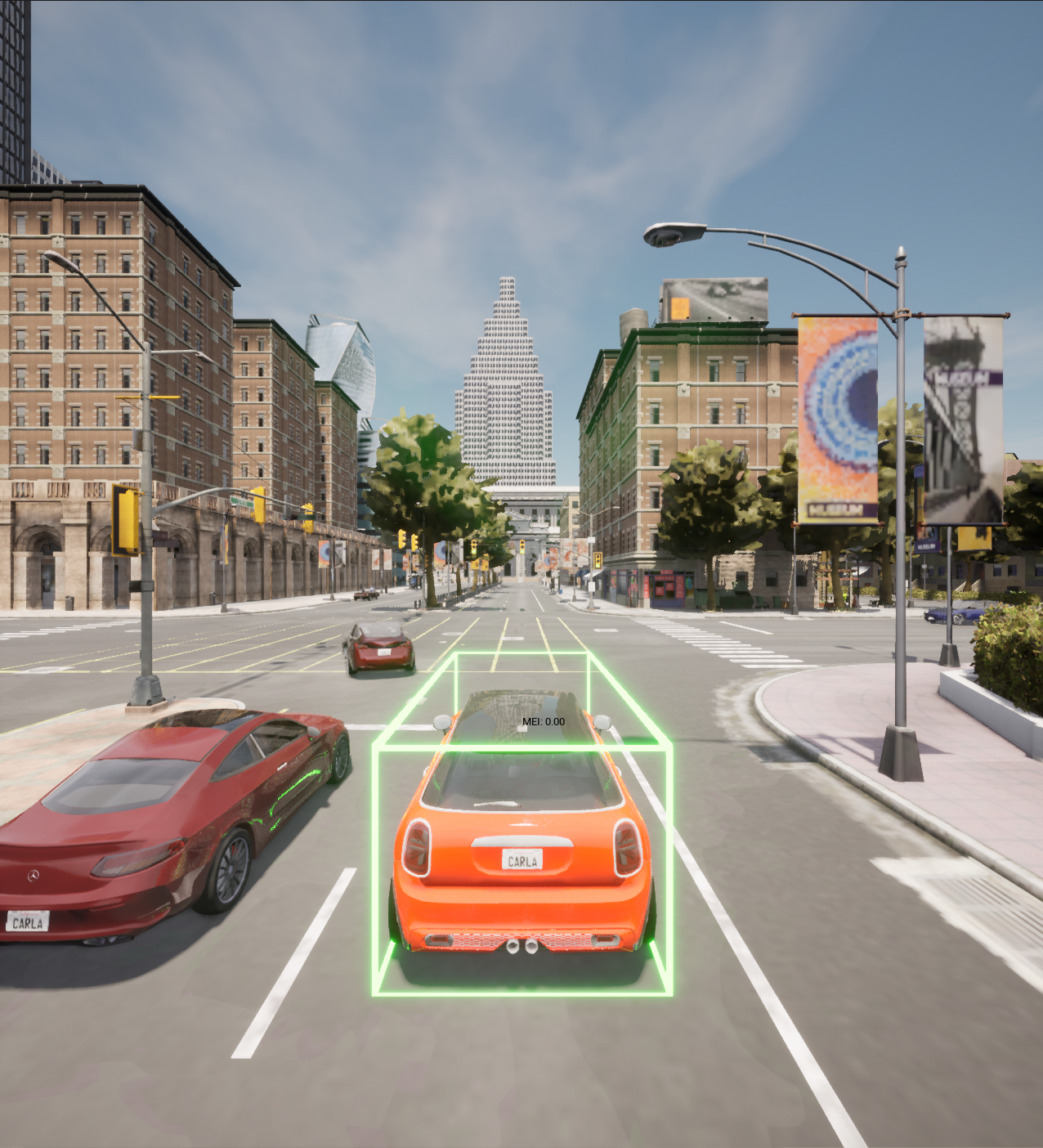}
        \caption{}
        \label{fig:scenario_presentation_sub1}
    \end{subfigure}
    \hfill
    \begin{subfigure}[b]{0.155\textwidth}
        \centering
        \includegraphics[width=\textwidth]{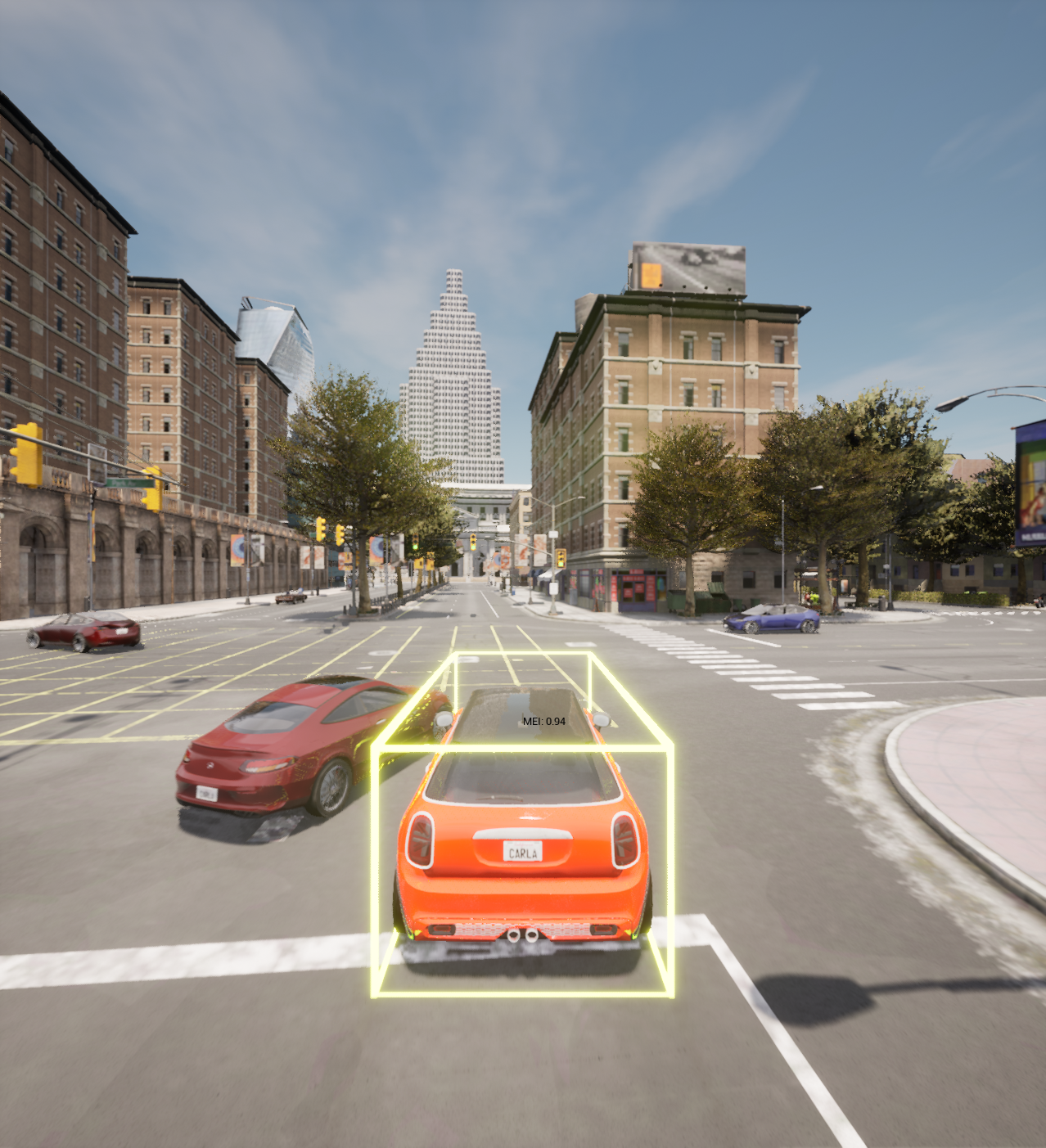}
        \caption{}
        \label{fig:scenario_presentation_sub2}
    \end{subfigure}
    \hfill
    \begin{subfigure}[b]{0.155\textwidth}
        \centering
        \includegraphics[width=\textwidth]{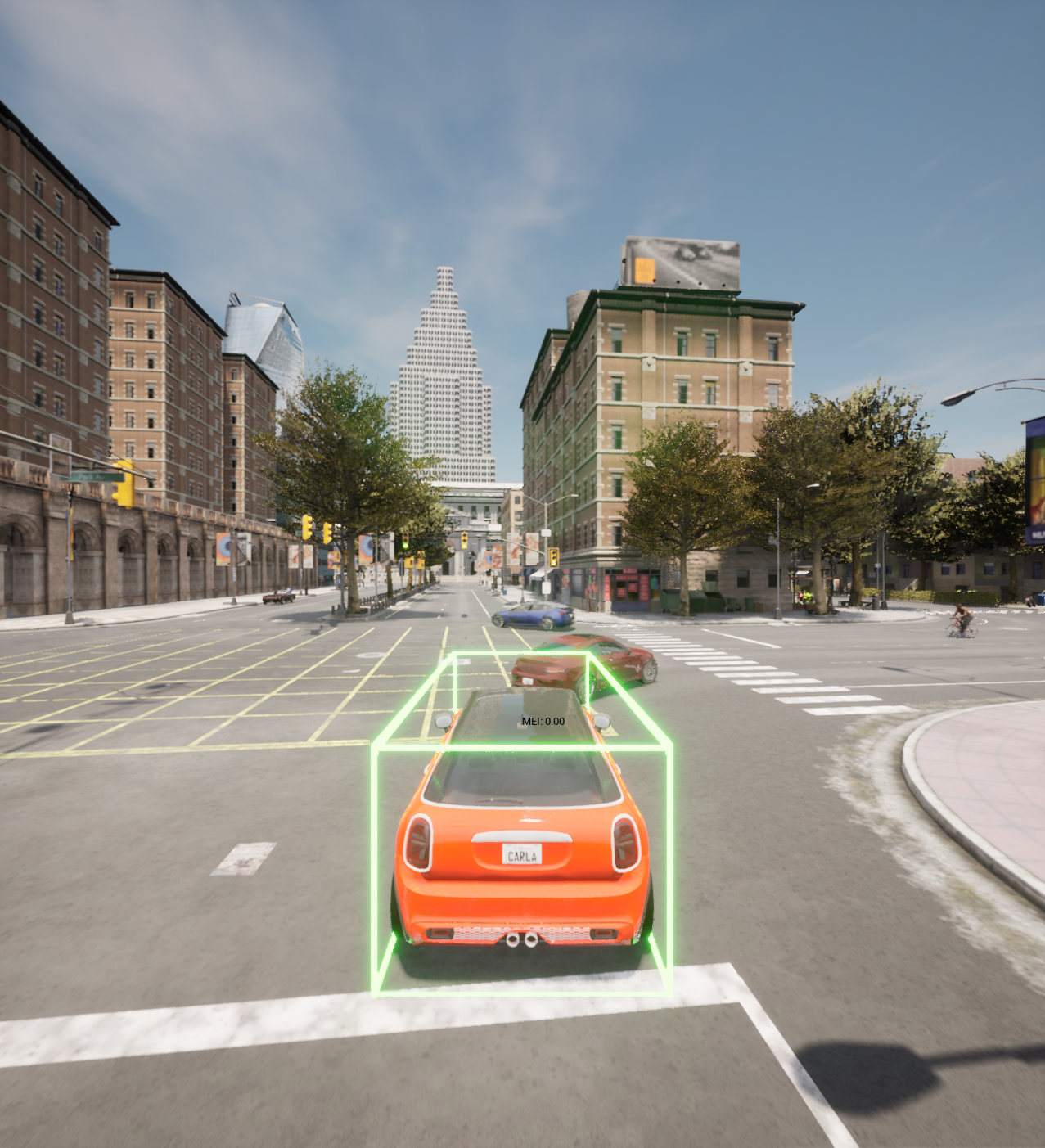}
        \caption{}
        \label{fig:scenario_presentation_sub3}
    \end{subfigure}
    
    \vspace{0.5em} 
    
    \begin{subfigure}[b]{0.155\textwidth}
        \centering
        \includegraphics[width=\textwidth]{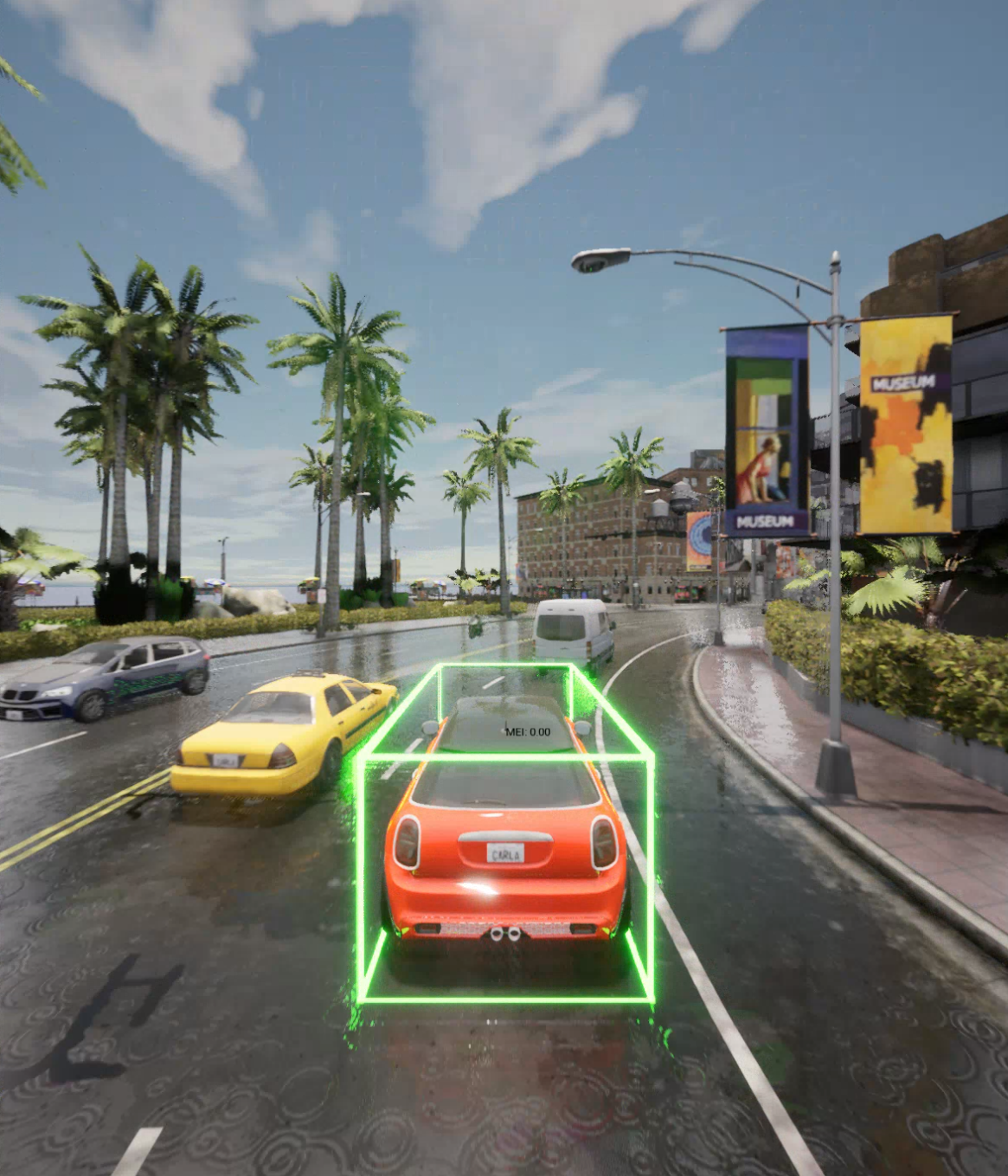}
        \caption{}
        \label{fig:scenario_presentation_sub4}
    \end{subfigure}
    \hfill
    \begin{subfigure}[b]{0.155\textwidth}
        \centering
        \includegraphics[width=\textwidth]{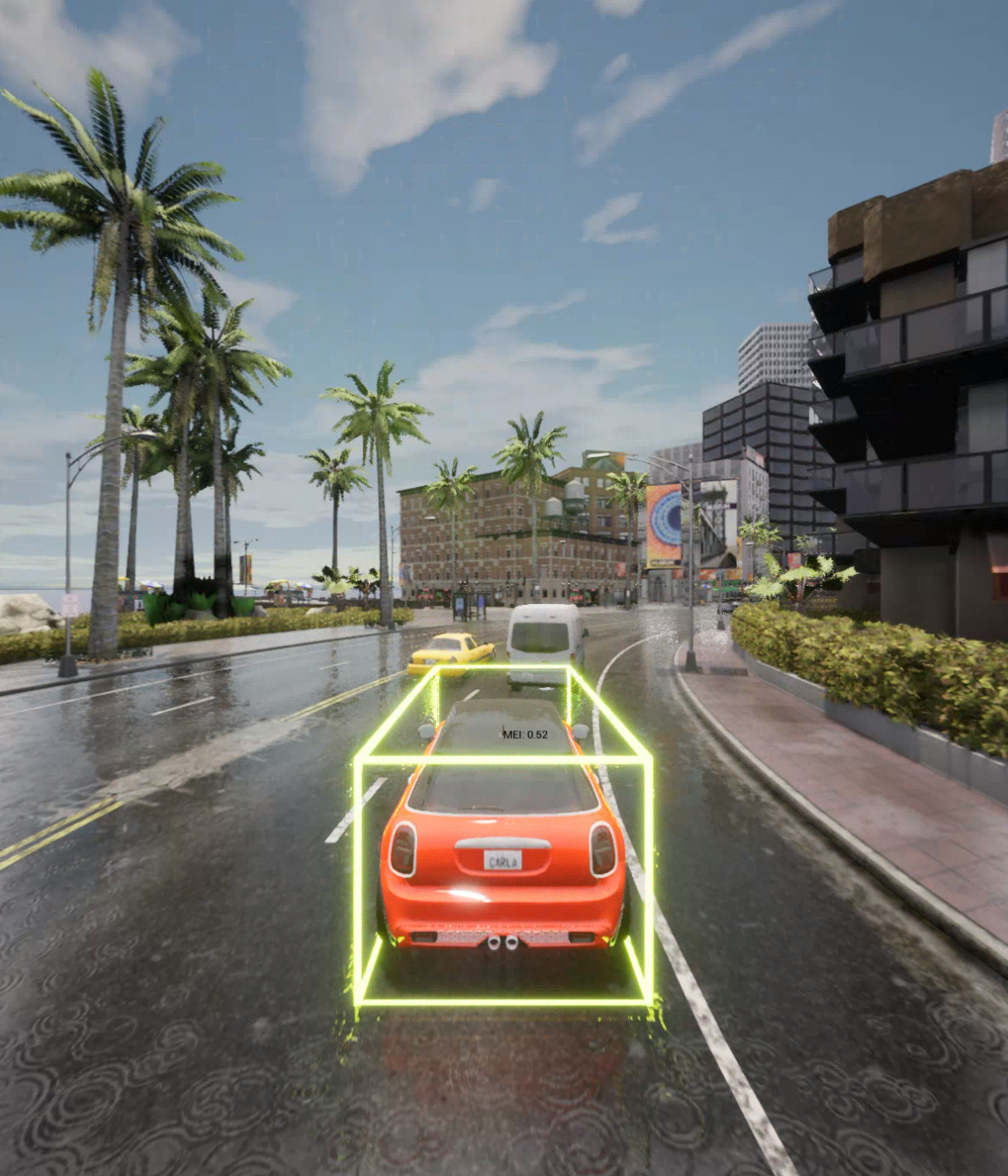}
        \caption{}
        \label{fig:scenario_presentation_sub5}
    \end{subfigure}
    \hfill
    \begin{subfigure}[b]{0.155\textwidth}
        \centering
        \includegraphics[width=\textwidth]{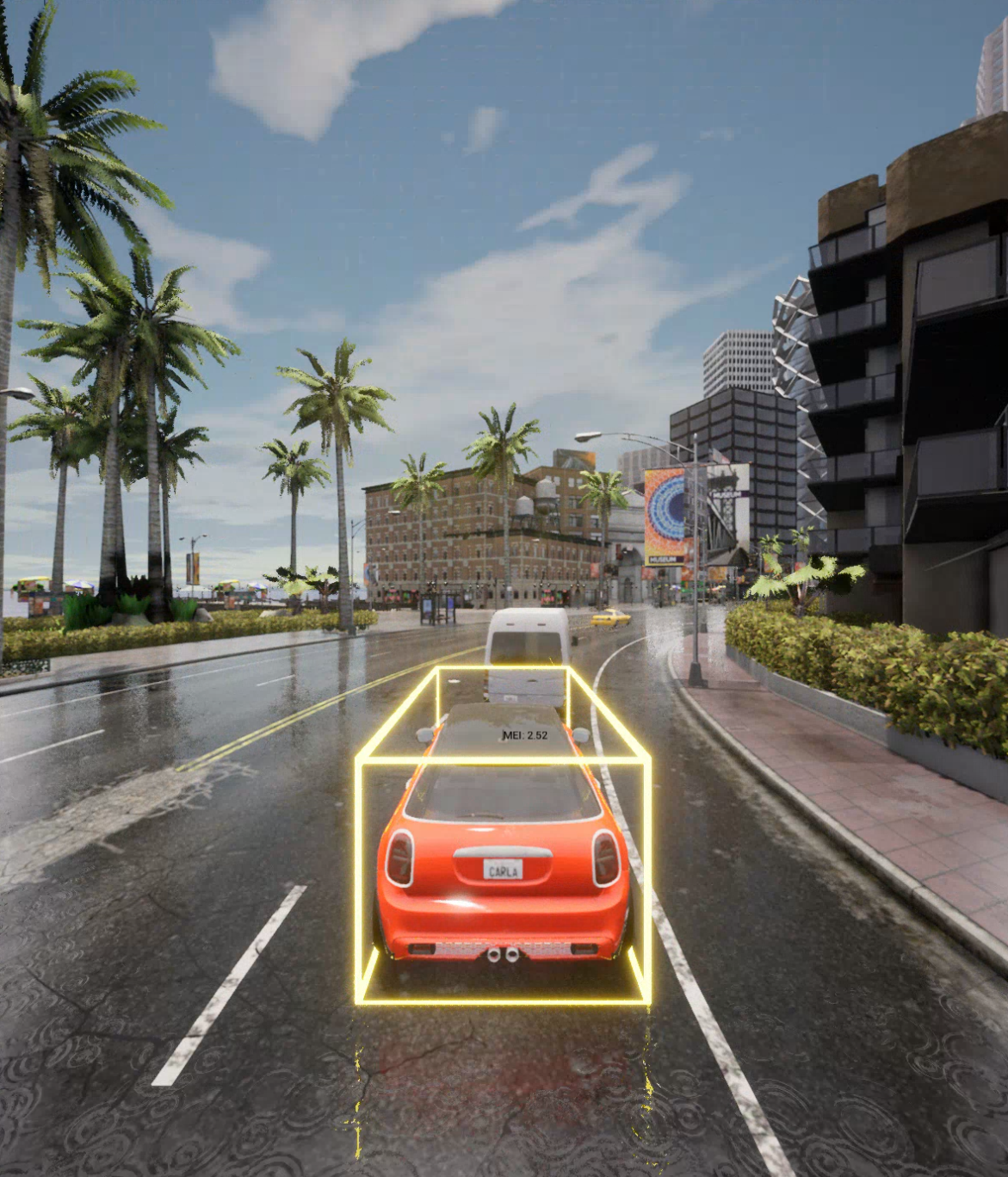}
        \caption{}
        \label{fig:scenario_presentation_sub6}
    \end{subfigure}

    \vspace{0.5em} 
    
    \begin{subfigure}[b]{0.155\textwidth}
        \centering
        \includegraphics[width=\textwidth]{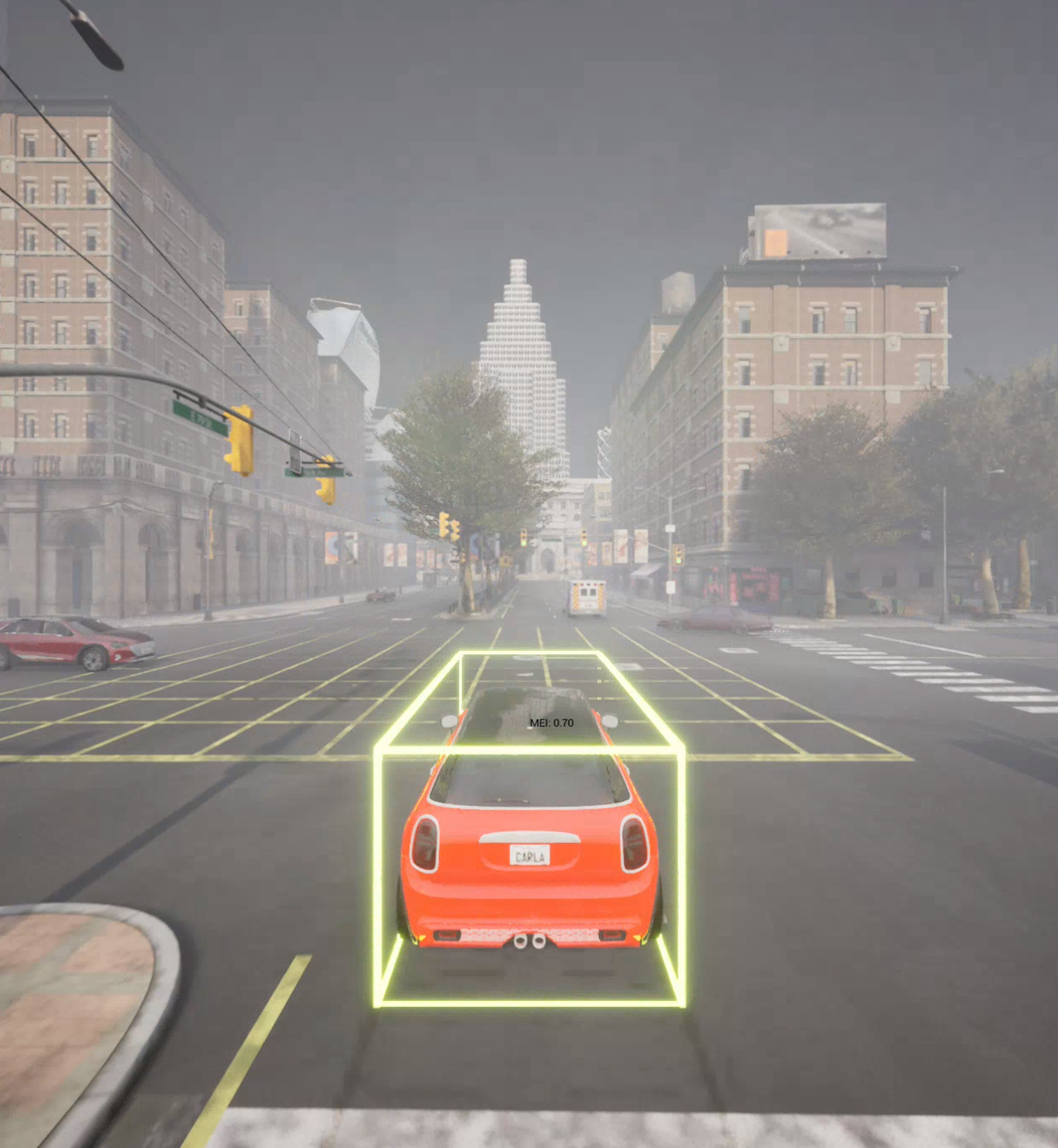}
        \caption{}
        \label{fig:scenario_presentation_sub7}
    \end{subfigure}
    \hfill
    \begin{subfigure}[b]{0.155\textwidth}
        \centering
        \includegraphics[width=\textwidth]{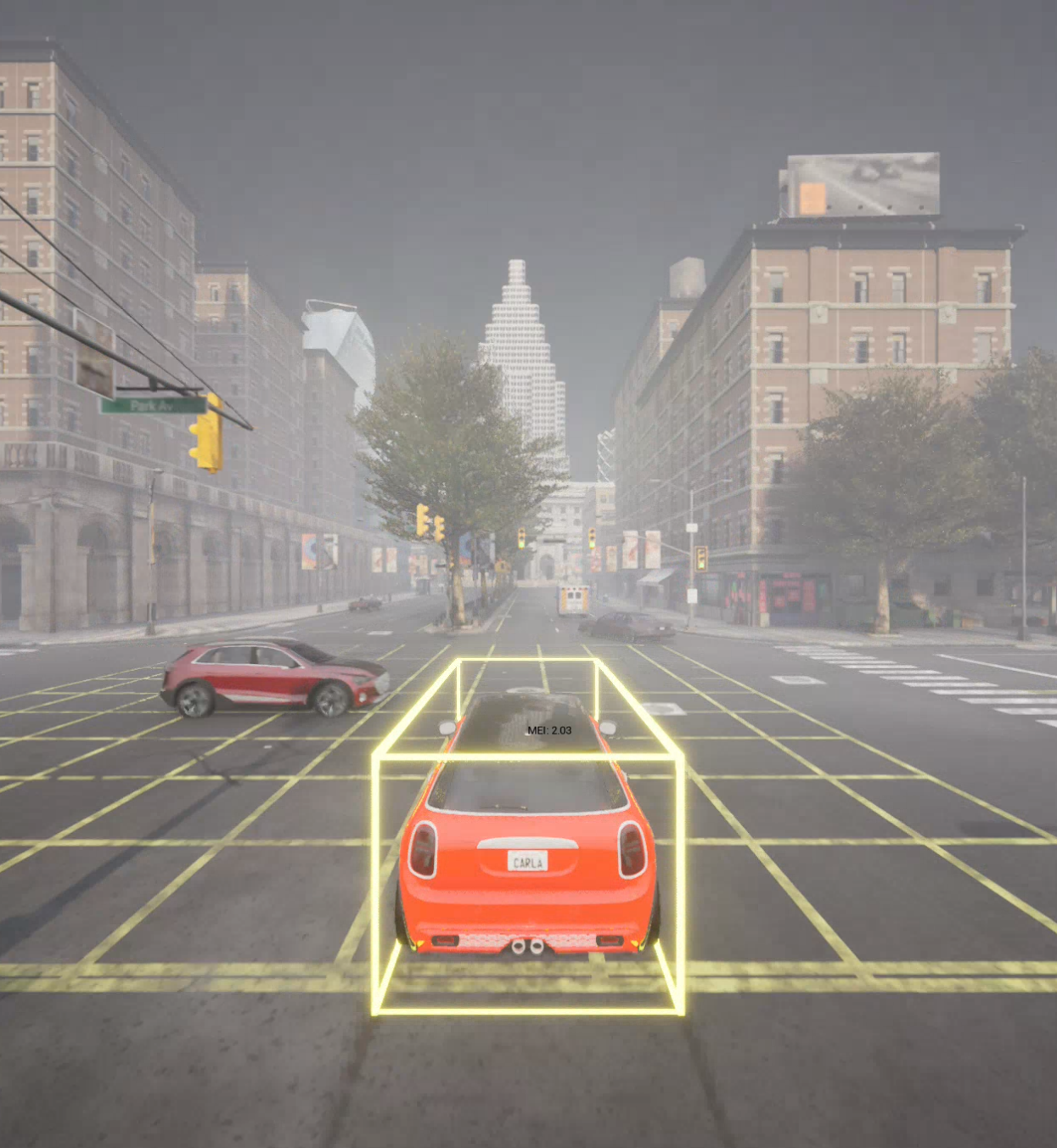}
        \caption{}
        \label{fig:scenario_presentation_sub8}
    \end{subfigure}
    \hfill
    \begin{subfigure}[b]{0.155\textwidth}
        \centering
        \includegraphics[width=\textwidth]{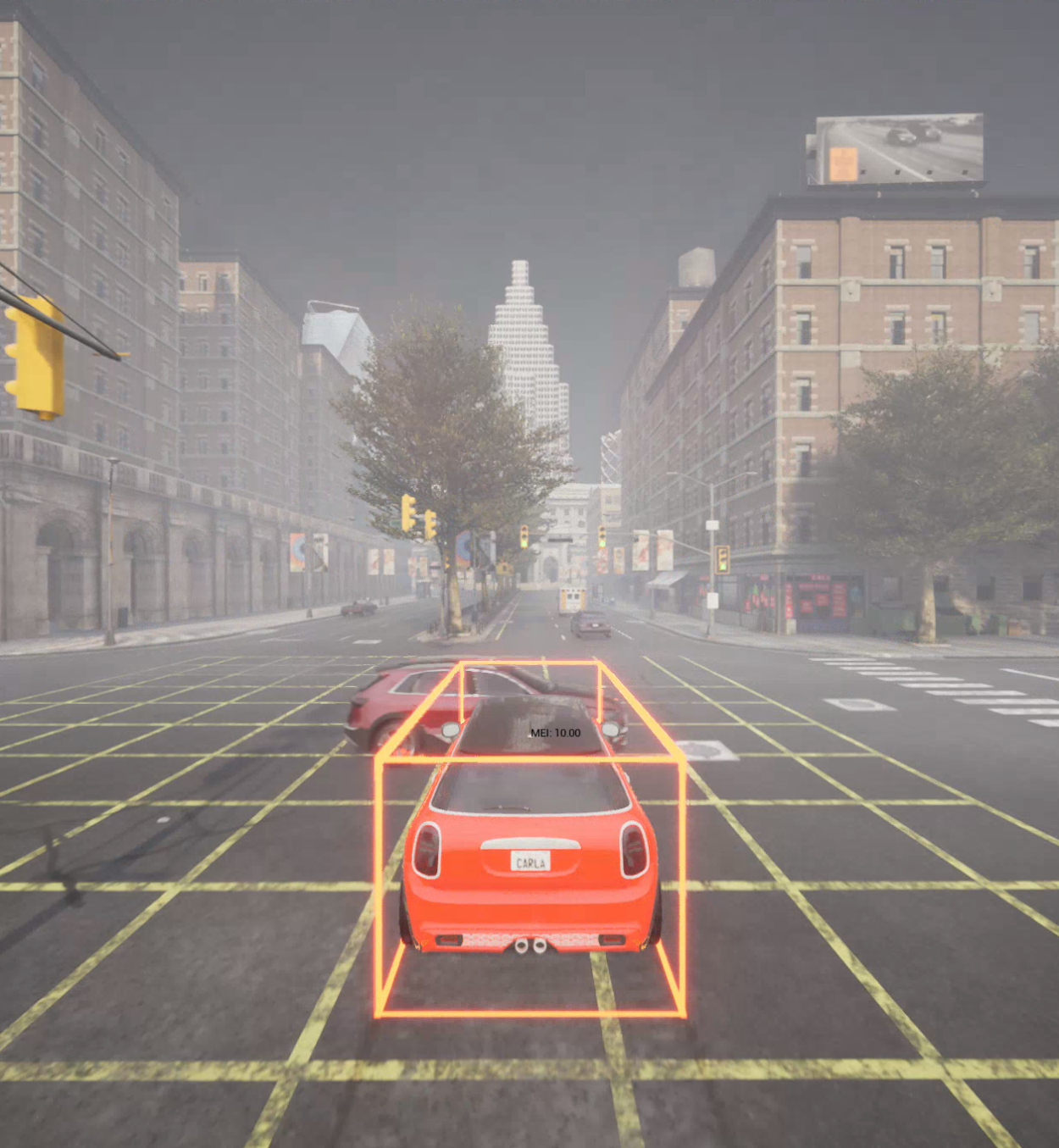}
        \caption{}
        \label{fig:scenario_presentation_sub9}
    \end{subfigure}
    
    \caption{Scenarios generated from different user descriptions. Figures \ref{fig:scenario_presentation_sub1}, \ref{fig:scenario_presentation_sub2}, and \ref{fig:scenario_presentation_sub3} represent a scenario generated based on \textit{Force Right Turn}, Figures \ref{fig:scenario_presentation_sub4}, \ref{fig:scenario_presentation_sub5}, and \ref{fig:scenario_presentation_sub6} represent a scenario generated based on \textit{Sudden Stop}, Figures \ref{fig:scenario_presentation_sub7}, \ref{fig:scenario_presentation_sub8}, and \ref{fig:scenario_presentation_sub9} represent a scenario generated based on \textit{Running Red Light}.}
    \label{fig:scenario_presentation}
\end{figure}

In this section, we present multiple interactive scenarios generated by \textit{LinguaSim}. The input descriptions selected are shown in Table \ref{tab:scenario_presentation}, and the corresponding generation results are presented in Fig. \ref{fig:scenario_presentation}. Figures \ref{fig:scenario_presentation_sub1}, \ref{fig:scenario_presentation_sub2}, and \ref{fig:scenario_presentation_sub3} represent one of the scenarios generated based on \textit{Force Right Turn}, where the red sedan performs a forced right turn, ignoring the traffic rules. Figures \ref{fig:scenario_presentation_sub4}, \ref{fig:scenario_presentation_sub5}, and \ref{fig:scenario_presentation_sub6} represent one of the scenarios generated based on \textit{Sudden Stop}, where the white van performs an emergency brake, and the ego vehicle manages to decelerate to avoid a collision. Figures \ref{fig:scenario_presentation_sub7}, \ref{fig:scenario_presentation_sub8}, and \ref{fig:scenario_presentation_sub9} represent one of the scenarios generated based on \textit{Running Red Light}, where the red sedan on the left runs the red light and almost gets hit by the ego vehicle.

\renewcommand{\arraystretch}{1.3}
\begin{table}[h]
\centering
\caption{Multiple input description used for \textit{LinguaSim} Scenario Generation}
\label{tab:scenario_presentation}
\begin{tabularx}{\columnwidth}{|>{\hsize=0.2\hsize}X|>{\hsize=0.8\hsize}X|}
    \hline
    \centering \textbf{\\Force \\ Right Turn} & “On a beautiful day, in front of a crossroad, a sedan on the left turns right without fully overtaking the ego, ignoring the lane marks."\\
    \hline
    \centering \textbf{\\Sudden \\ Stop} & “On a rainy day, on a curved road, a van is driving ahead of the ego vehicle at a moderate speed. A while later, it suddenly brakes at maximum deceleration and remains idle"\\
    \hline
    \centering \textbf{\\Running \\ Red Light} & “On a misty morning, in front of a crossroad,  there's almost no one on the street, a sedan at the left entrance of the intersection ignores the red light and drives through the intersection, with no intention to yield to the ego vehicle"\\
    \hline
\end{tabularx}
\end{table}

\subsection{Scenario Generation with Different Criticality}
\label{sec:scenario_generation}
In this section, we incorporate real-time safety evaluation metrics into the generated scenarios and conduct a quantitative analysis to assess their criticality and comfort. By comparing the performance of these scenarios across various metrics, we aim to evaluate how different descriptions influence the scenario outcomes.

\subsubsection{Setup}
We use the descriptions in Table \ref{tab:scenario_setup} as examples to compare the average metrics of the generated scenarios. For this comparison, the \textit{Refine Commander} and \textit{Refiner} agents are not activated. We generate 32 different scenarios based on all three descriptions and calculate the mean value of each metric to provide an overall comparison.

\begin{table}[h]
\centering
\caption{Comparison between Scenario Batches Generated Based on Different Natural Language Inputs}
\label{tab:scenario_setup}
\begin{tabularx}{\columnwidth}{|>{\hsize=0.2\hsize}X|>{\hsize=0.8\hsize}X|}
    \hline
    \centering \textbf{\\Dangerous \\ Description} & "In a heavy traffic, a truck on the left overtakes the ego vehicle, cut in aggressively and suddenly brakes, then kept going, a sedan follows the ego vehicle very closely behind."\\
    \hline
    \centering \textbf{Moderate \\ Description} & "A truck on the left overtakes the ego vehicle, cut in, then kept going, a sedan follows the ego vehicle."\\
    \hline
    \centering \textbf{\\Safe \\ Description} & "There's almost no one on the street, pretty quiet, a truck on the left overtakes the ego vehicle, cuts in in a decent way, then keeps going, a sedan follows the ego vehicle remotely."\\
    \hline
\end{tabularx}
\end{table}

\smallskip
\subsubsection{Metrics}
To quantify the differences between scenarios generated from various descriptions, we employ three distinct metrics that evaluate different aspects of scenario performance:
\begin{itemize}
    \item \textit{Anticipated Collision Time ($ACT$)} \cite{b12}, a surrogate safety measure that is widely used for real-time safety evaluation  \cite{b13} \cite{b14} \cite{b15}, proposed by Venthuruthiyil and Chunchu, which is calculated according to Equation \ref{eq:act} frame by frame throughout the scenario:
    \begin{equation}
        \label{eq:act}
        \text{ACT} = 
        \begin{cases} 
        \frac{\delta}{\frac{d\delta}{dt}}, & \text{if } \frac{d\delta}{dt} > 0 \\
        +\infty, & \text{otherwise}
        \end{cases}
    \end{equation}
    where $\delta$ is the shortest distance between the vehicles. We use the minimum value across the entire timeline to represent the overall criticality of the scenario (as lower ACT values indicate more critical situations with less time to potential collision).

    \smallskip
    
    \item \textit{Comfortability}, first calculated for each frame according to Equation \ref{eq:comfortability}:
        \begin{equation}
            \label{eq:comfortability}
            \mathcal{C} = \mathbb{E}\left[\frac{1}{\|\tilde{\mathbf{a}}_t\| + 1} \,\bigg|\|\tilde{\mathbf{a}}_t\| \neq 0\right]
        \end{equation}
    where $\tilde{\mathbf{a}}_t$ represents the denoised acceleration in $m/s^2$, and then the mean value of these calculations throughout the entire scenario is used to assess the general comfort level during the time span. Comfortability is expressed as a score ranging from 0 to 1, with higher values indicating better comfort. 

    \smallskip
    
    \item  \textit{Crash Rate ($CR$)}, quantifies the crash probability and is defined as the ratio of scenarios with collisions to the total number of simulated scenarios.
\end{itemize}
Together, these metrics provide a comprehensive evaluation framework for comparing scenarios generated from different descriptions.

\begin{figure}[h]
    \centering
    \includegraphics[width=1\linewidth]{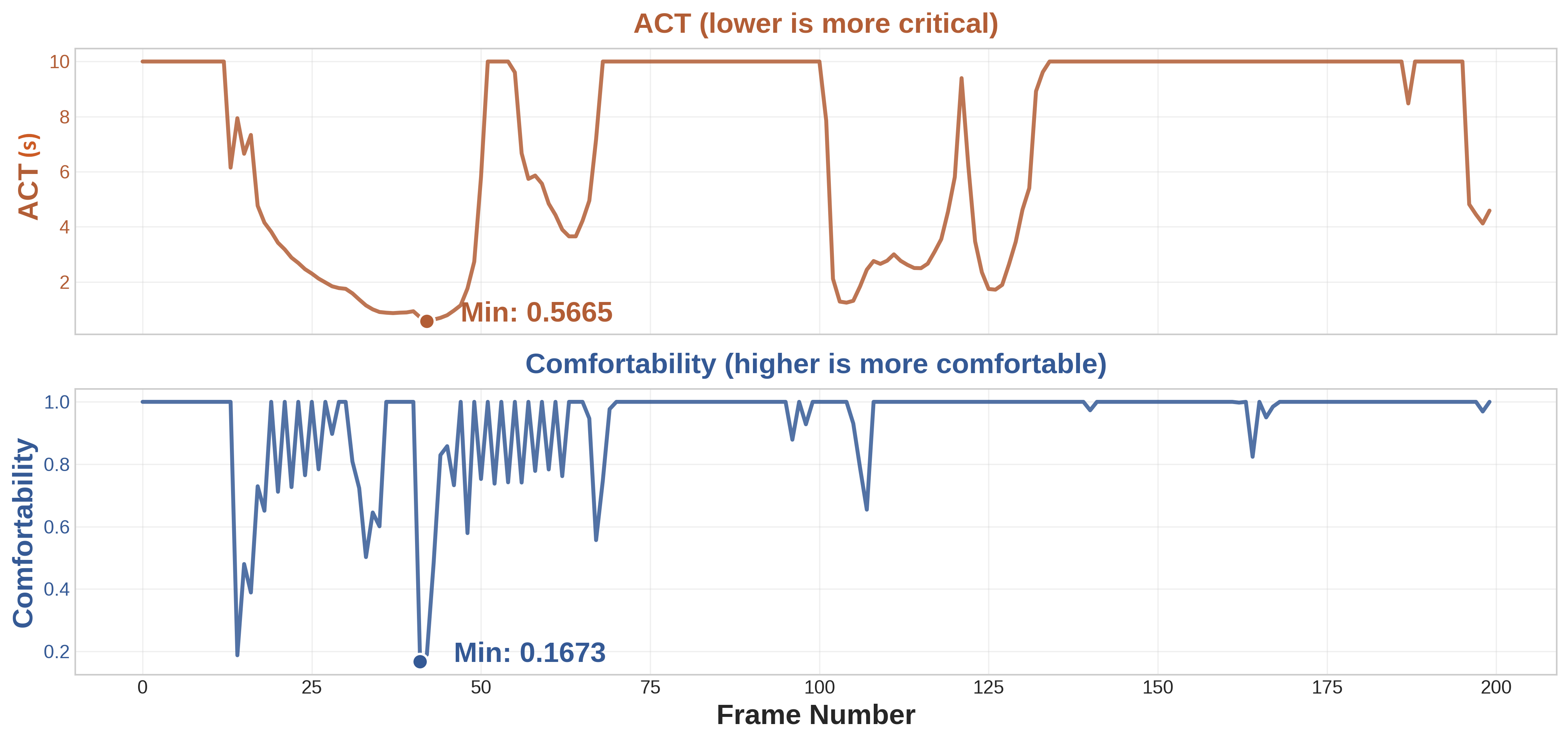}
    \caption{\textit{ACT} and \textit{Comfortability} curve in a single generated scenario, where the critical point shows the moment where the truck cut in in front of the ego vehicle and suddenly brake, forcing it to execute an emergency brake}
    \label{fig:metrics}
\end{figure}

\subsubsection{Interactive Scenarios Generation}
\begin{figure}[htbp]
    \centering
    \begin{subfigure}[b]{0.155\textwidth}
        \centering
        \includegraphics[width=\textwidth]{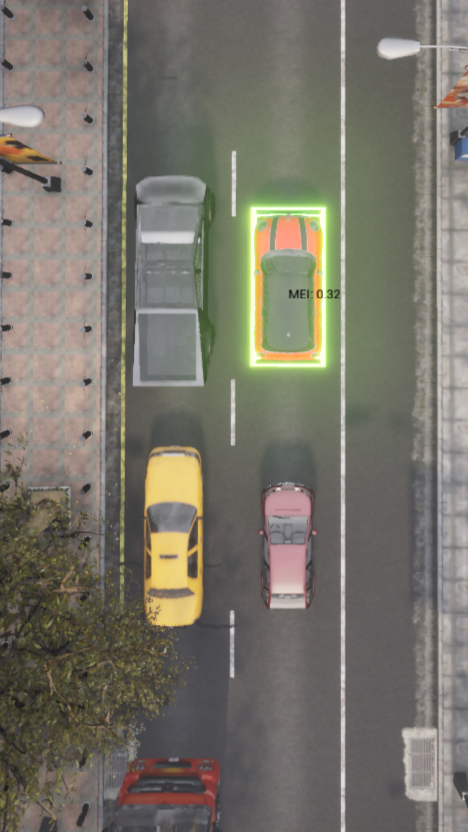}
        \caption{}
        \label{fig:sub1}
    \end{subfigure}
    \hfill
    \begin{subfigure}[b]{0.155\textwidth}
        \centering
        \includegraphics[width=\textwidth]{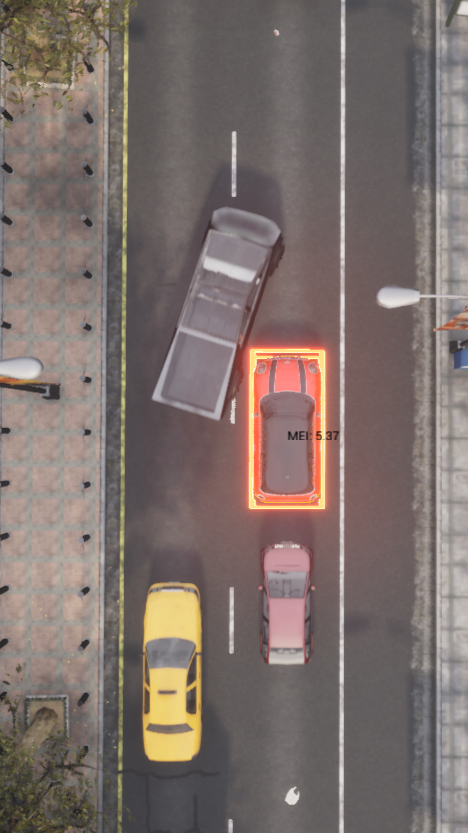}
        \caption{}
        \label{fig:sub2}
    \end{subfigure}
    \hfill
    \begin{subfigure}[b]{0.155\textwidth}
        \centering
        \includegraphics[width=\textwidth]{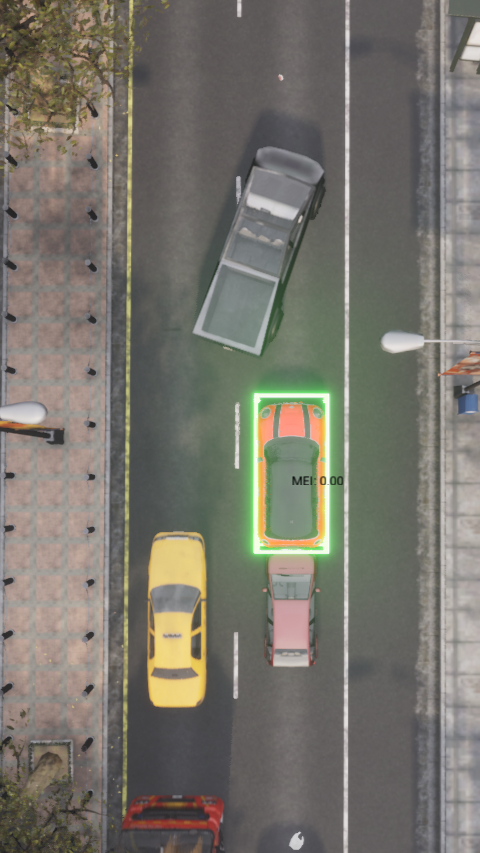}
        \caption{}
        \label{fig:sub3}
    \end{subfigure}
    
    \vspace{0.5em} 
    
    \begin{subfigure}[b]{0.155\textwidth}
        \centering
        \includegraphics[width=\textwidth]{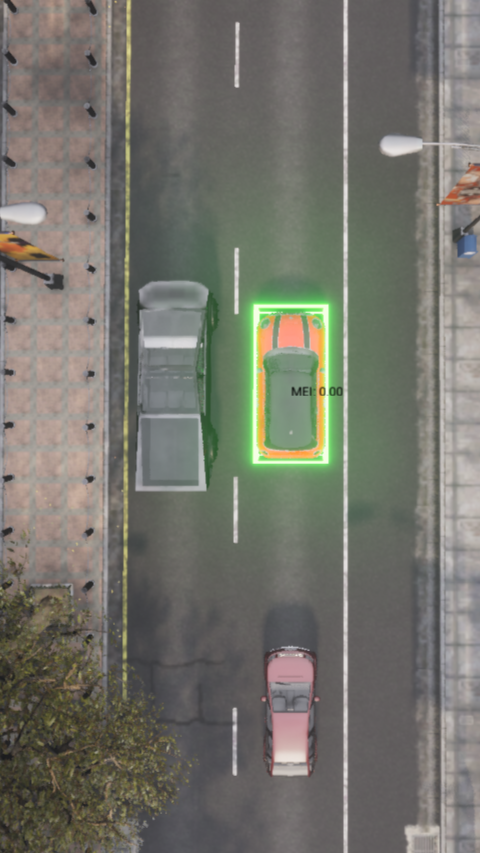}
        \caption{}
        \label{fig:sub4}
    \end{subfigure}
    \hfill
    \begin{subfigure}[b]{0.155\textwidth}
        \centering
        \includegraphics[width=\textwidth]{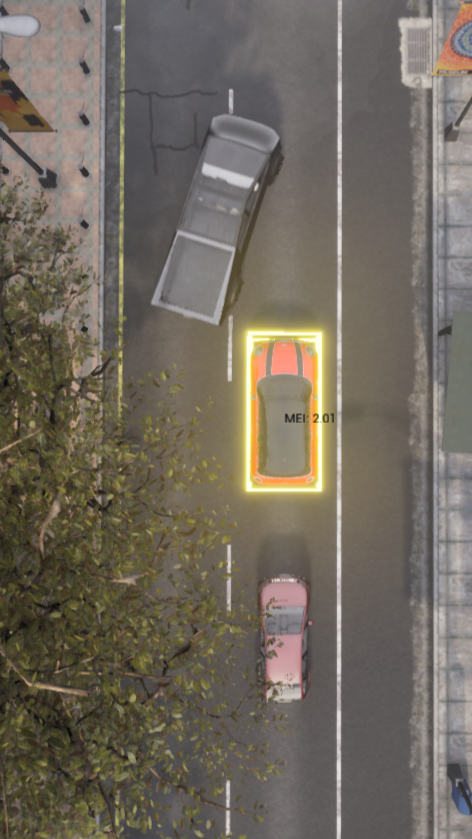}
        \caption{}
        \label{fig:sub5}
    \end{subfigure}
    \hfill
    \begin{subfigure}[b]{0.155\textwidth}
        \centering
        \includegraphics[width=\textwidth]{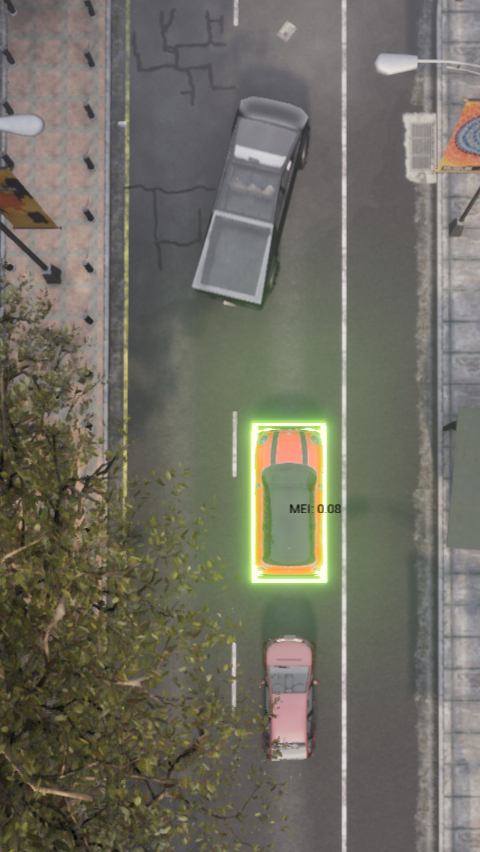}
        \caption{}
        \label{fig:sub6}
    \end{subfigure}
    
    \caption{The scenario generated by different user descriptions, Fig. \ref{fig:sub1}, \ref{fig:sub2}, \ref{fig:sub3} represent one of the scenario generated based on \textit{Dangerous Description}, where Fig. \ref{fig:sub2}, \ref{fig:sub3} show that the truck forces a cut-in, causing the ego vehicle to perform a sudden brake, eventually lead to a collision between the ego vehicle and the sedan behind it. Fig. \ref{fig:sub4}, \ref{fig:sub5}, \ref{fig:sub6} represent one of the scenario generated based on \textit{Safe Description}, where Fig. \ref{fig:sub5} shows that the truck only starts to cut in after a clean overtake, while the sedan follows in a safe distance.}
    \label{fig:cut-in}
\end{figure}

The scenarios generated by \textit{LinguaSim} are shown in Fig. \ref{fig:cut-in}. Due to the space constraints, only the results generated from \textit{Dangerous Description} and \textit{Safe Description} are displayed.

\smallskip

Table \ref{tab:scenario_comparison} highlights the differences in criticality and comfortability among the three batches of scenarios generated from distinct descriptions. Notably, when users describe safer traffic scenes, we observe a significant increase in \textit{ACT}, which indicates reduced criticality. Additionally, comfortability increases as the criticality of the traffic scene decreases, primarily due to less aggressive following and cut-in behaviors. In contrast, scenarios generated from the \textit{Dangerous Description} exhibit the highest criticality, with a \textit{Crash Rate} of 46.9\%; although this outcome does align with the aggressive traffic description, it still contradicts the user's description of "almost hit". Our proposed solution to address this discrepancy is detailed in Section \ref{subsec:refinement}.

\renewcommand{\arraystretch}{1.3}
\begin{table}[h]
\centering
\caption{Comparison between Scenario Batches Generated Based on Different Natural Language Inputs}
\label{tab:scenario_comparison}
\begin{tabular}{|c|c|c|c|c|}
    \hline
    \diagbox{Scenarios}{Metric} & ACT & Comfortability & CR\\
    \hline
    Dangerous Description & 0.072 & 0.654 & 46.9\%\\
    \hline
    Moderate Description & 0.938 & 0.722 & 0.0\%\\
    \hline
    Safe Description & 3.532 & 0.764 & 0.0\%\\
    \hline
\end{tabular}
\end{table}

\subsection{Scenario Refinement}
\label{subsec:refinement}
In this section, we focus on refining the previously generated scenarios in section \ref{sec:scenario_generation}, as detailed in Table \ref{tab:scenario_comparison}. When using the \textit{Dangerous Description} to generate scenarios, despite the user specifying "\textit{A sedan that follows the ego vehicle very closely behind almost hits the ego}", \textit{LinguaSim} occasionally produces overly aggressive scenarios, resulting in a 46.9\% \textit{Crash Rate}, which deviates from the user's original intention. To address this, the \textit{Refine Commander} and \textit{Refiner} agents are activated. All 32 scenarios previously generated from the \textit{Dangerous Description} in Section \ref{sec:scenario_generation} are refined to ensure precise results. The \textit{Refiner} agent performs up to five refinement episodes for each scenario.

\begin{table}[h]
\centering
\caption{Comparison Between the Original Scenarios and the Refined Scenarios Generated Based on \textit{Dangerous Description}}
\label{tab:refinement_result}
\begin{tabular}{|c|c|c|c|}
    \hline
    \diagbox{Scenarios}{Metric} & ACT & Comfortability & CR\\
    \hline
    Original Scenarios & 0.072 & 0.654 & 46.9\%\\
    \hline
    Refined Scenarios & 0.214 & 0.691 & 6.3\%\\
    \hline
\end{tabular}
\end{table}

The results of the refinement process are summarized in Table \ref{tab:refinement_result}. Following refinement, the \textit{Anticipated Collision Time (ACT)} increases from 0.072 s to 0.214 s. These changes indicate a reduction in overall criticality while maintaining a level of danger compared to the moderate and safe scenarios described in Section \ref{sec:scenario_generation}. Notably, the \textit{Crash Rate (CR)} drops substantially from $46.9\%$ to $6.3\%$, better aligning with the user's intention of creating a dangerous yet non-collision scenario.

\section{Conclusion}
\label{sec:conclusion}
In this paper, we presented \textit{LinguaSim}, a framework that bridged the gap between natural language instructions and interactive multi-vehicle testing scenarios, with adversarial vehicle behaviors constrained through the combined use of scenario descriptions and autonomous driving models guiding them. Our layered generation structure ensured both accurate interpretation of user intent and realistic simulated vehicle behaviors, while the feedback calibration module refined scenario precision by correcting overly aggressive interactions. Experimental results demonstrated \textit{LinguaSim}’s effectiveness in producing scenarios with criticality levels aligned to different language descriptions, with Anticipated Collision Time (ACT) ranging from 0.072 s for dangerous to 3.532 s for safe scenarios, and comfortability scores increasing from 0.654 to 0.764 accordingly. The refinement process notably reduced the crash rate from 46.9\% to 6.3\%, better matching user intentions while maintaining realistic driving behaviors. Future work could extend this research by training autonomous driving models directly on our generated scenarios to further validate their realism and utility. Additionally, expanding the diversity of generated scenarios could better adapt the framework to various training and testing situations.

\medskip

While LinguaSim demonstrates promising capabilities in natural language-driven scenario generation, the current experimental validation is preliminary. Further evaluation against established benchmarks and broader testing across diverse simulation environments would strengthen the assessment of scalability and reliability for safety-critical applications. Future research will focus on conducting large-scale comparative studies with state-of-the-art methods, implementing robust validation frameworks, and developing comprehensive failure analysis protocols to enhance the reliability and practical applicability of LinguaSim in autonomous vehicle testing and training scenarios.

\vspace{12pt}

\end{document}